\documentclass{article}
\usepackage{spconf}

\usepackage{amsmath,amssymb,amsthm,amsbsy}
\usepackage{algorithm,algorithmic}
\usepackage{graphicx,subcaption}
\usepackage{booktabs,multirow,diagbox}
\usepackage{enumitem}
\usepackage{cite}
\usepackage{url}

\theoremstyle{remark}

\newcommand{\expect}{\mathbb{E}}

\newcommand{\KL}{\mathcal{D}}
\newcommand{\CE}{\mathcal{H}}

\title{Zero-Shot Learning of a Conditional Generative Adversarial Network for Data-Free Network Quantization}
\name{Yoojin Choi, Mostafa El-Khamy, Jungwon Lee}
\address{SoC R\&D, Samsung Semiconductor Inc., San Diego, CA 92121, USA}
\begin{document}
%
\maketitle
\begin{abstract}
We propose a novel method for training a conditional generative adversarial network (CGAN) without the use of training data, called zero-shot learning of a CGAN (ZS-CGAN). Zero-shot learning of a conditional generator only needs a pre-trained discriminative (classification) model and does not need any training data. In particular, the conditional generator is trained to produce labeled synthetic samples whose characteristics mimic the original training data by using the statistics stored in the batch normalization layers of the pre-trained model. We show the usefulness of ZS-CGAN in data-free quantization of deep neural networks. We achieved the state-of-the-art data-free network quantization of the ResNet and MobileNet classification models trained on the ImageNet dataset. Data-free quantization using ZS-CGAN showed a minimal loss in accuracy compared to that obtained by conventional data-dependent quantization.
\end{abstract}

\begin{keywords}
Zero-shot learning, conditional generative adversarial networks, data-free training, quantization
\end{keywords}

\section{INTRODUCTION}\label{sec:intro}

Generative adversarial networks (GANs)~\cite{goodfellow2014generative} are of great interest in deep learning for image or speech synthesis problems. Two neural networks, called generator and discriminator, play a zero-sum game to learn the mapping from a random noise distribution to the target data distribution. The generator is trained to fool the discriminator by making its fake samples as similar as possible to the real training data, while the discriminator is trained to distinguish the fake samples produced by the generator from the real training data. Conditional GANs (CGANs)~\cite{mirza2014conditional} are the conditional version of GANs, where both the generator and the discriminator are conditioned on some extra information, such as classes or attributes. CGANs have potential in various conditional generation tasks such as labeled image generation, image-to-image translation, and so on~\cite{miyato2018cgans}. In this paper, we propose a novel method of training a conditional GAN without any training data, called zero-shot learning of a CGAN (ZS-CGAN).

Network quantization is an important procedure for efficient inference when deploying pre-trained deep neural networks on resource-limited platforms~\cite{deng2020model}. By using quantized weights and activations, we not only reduce the computational cost, but also lower the memory footprint required for inference. The improved efficiency after quantization is usually traded off against an accuracy loss. To minimize the accuracy loss from quantization, the quantization parameters are optimized with some calibration data in post-training quantization (PTQ). To recover the accuracy loss, the quantized model can also be re-trained with some training data, which is called quantization-aware training (QAT).

Most of the existing network quantization methods are data-dependent, implying that a large number of training data, which were already used in training the floating-point model, are assumed to be available and used in the quantization procedure (cf. \cite{rastegari2016xnor,zhou2016dorefa,choi2017towards,jacob2018quantization,zhang2018lq,wang2019haq}). However, it becomes more and more difficult and expensive to share the training data due to their large size, restriction by proprietary rights, and to preserve data privacy. The regulations and compliance requirements around privacy and security complicate both data sharing by the original model trainer and data collection by the model quantizer, for example, in the case of medical and bio-metric data.

\begin{figure}[t]
\centering
\includegraphics[width=\columnwidth]{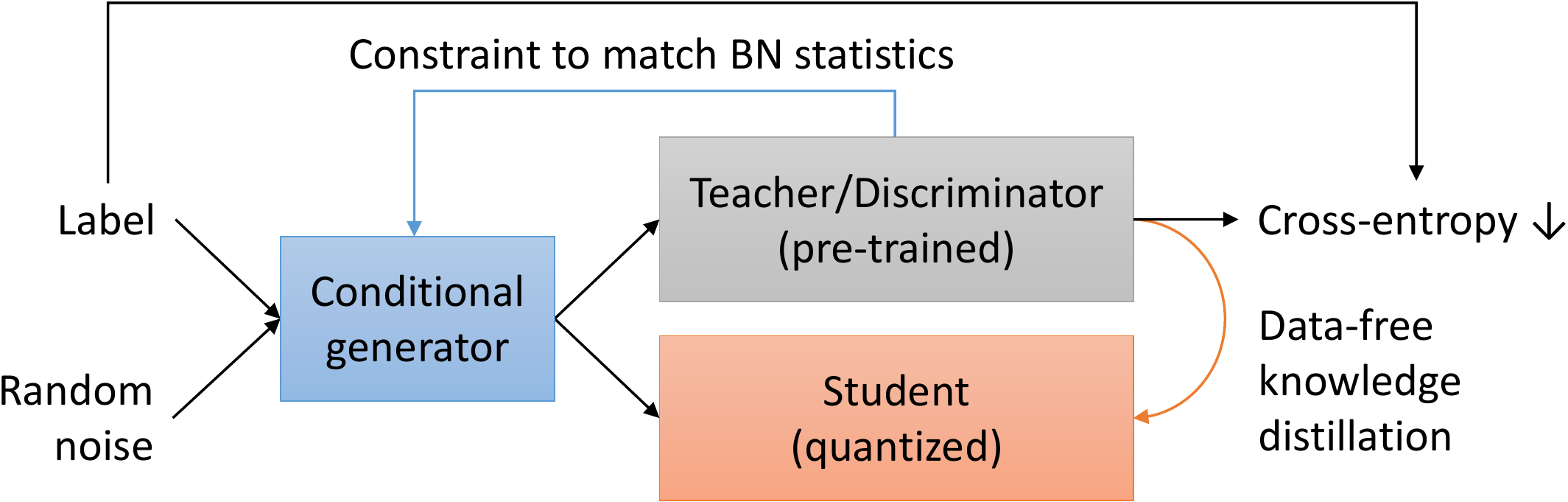}
\caption{Zero-shot learning of a CGAN (ZS-CGAN) and data-free knowledge distillation. First, we train a conditional generator without any training data. A pre-trained classification model (called teacher) plays the role of a (fixed) discriminator that evaluates generated samples. In particular, the cross-entropy between the generator input and the teacher output is minimized. The generator is also constrained to produce synthetic samples similar to the original training data by matching the statistics at the batch normalization (BN) layers of the teacher. Second, we transfer knowledge from the pre-trained teacher to a (quantized) student via data-free knowledge distillation using the synthetic samples from the generator.\label{sec:intro:fig:01}}\vspace{-1em}
\end{figure}


The conditional generator trained with our proposed ZS-CGAN can generate samples whose statistics match those of the training dataset. Hence, we propose utilizing it for data-free network quantization. Data-free network quantization recently attracted interest to avoid complications of data sharing. Weight equalization and bias correction are proposed for data-free weight quantization in \cite{nagel2019data}. However, no synthetic data are produced or used for network quantization in \cite{nagel2019data}. In \cite{cai2020zeroq,yin2020dreaming,choi2020data,haroush2020knowledge}, it is proposed to produce synthetic data to (re-)train pruned or quantized models, when no original training data are accessible. In \cite{cai2020zeroq,yin2020dreaming,haroush2020knowledge}, synthetic samples are inferred directly in the image domain based on the statistics stored in the batch normalization layers of a pre-trained model. Non-conditional generators were introduced to generate synthetic samples that mimic the original data for data-free quantization in \cite{choi2020data}. Our experimental results show that conditionally generated samples using ZS-CGAN proposed in this paper are more diverse and realistic than non-conditionally generated samples, hence providing better quantization performance. 

Our proposed ZS-CGAN framework is demonstrated in Figure~\ref{sec:intro:fig:01}. Note that conventional GAN or CGAN training requires training data for the discriminator in its minimax optimization. However, we assume that no training data are available, and we are given a pre-trained classification model, instead. We use the pre-trained model as a fixed discriminator to criticize the synthetic samples from a conditional generator. The conditional generator takes a label as its input so it synthesizes labeled samples. We minimize the cross-entropy loss between the label fed to the generator and the softmax output from the pre-trained model for the generated sample. The statistics stored in batch normalization layers of the pre-trained model are additionally used to make the conditional generator to produce labeled synthetic samples similar to the original training data.

We evaluated the proposed scheme for data-free network quantization of ResNet~\cite{he2016deep} and MobileNet~\cite{howard2017mobilenets,sandler2018mobilenetv2,howard2019searching} models pre-trained on the ImageNet ILSVRC 2012 dataset~\cite{russakovsky2015imagenet}. With ZS-CGAN, we achieve the state-of-the-art data-free network quantization results for $8$-bit and $6$-bit quantization of both weights and activations. The proposed data-free network quantization scheme also shows very marginal accuracy degradation, compared to the conventional data-dependent quantization method that uses the original ImageNet dataset.

\section{DATA-FREE NETWORK QUANTIZATION}\label{sec:dfq}

\subsection{Conditional GANs (CGANs)}\label{sec:dfq:cgan}

Generative adversarial networks (GANs) were introduced in \cite{goodfellow2014generative} as a framework to learn generative models that can mimic the target data distribution. A GAN consists of two adversarial models: a generator~$G$ that captures the data distribution, and a discriminator~$D$ that estimates the probability whether a sample comes from the training data or is generated by $G$. A conditional GAN (CGAN) is a type of GANs that uses conditional information~$y$ for the generator and the discriminator. We can perform the conditioning by feeding $y$ into both the discriminator and the generator as additional input. The condition can be any information observed in the data, such as classes, attributes, or modalities. The two-player minimax game for training a CGAN is given by
\begin{multline}\label{sec:dfq:cgan:01}
\min_G\max_D
\{\expect_{p(x,y)}[\log D(x,y)] \\
+\expect_{p(z)p(y)}[\log(1-D(G(z,y),y)]\},
\end{multline}
where $p(x,y)$ is the target data distribution, $p(z)$ is a random noise distribution, and $p(y)$ is the distribution for condition~$y$.

\subsection{Zero-Shot Learning of a CGAN (ZS-CGAN)}\label{sec:dfq:zs-cgan}

As shown in \eqref{sec:dfq:cgan:01}, the original CGAN training was developed under the assumption that training data are given for the expectation over $p(x,y)$. Sharing a large training dataset is expensive and sometimes not even possible due to privacy and regulatory concerns. In this subsection, we consider data-free training of CGANs that we can use in the situation where any training data are not accessible.

We are given a pre-trained classification model~$T$ (called teacher) that estimates the probability distribution of class~$y$ for~$x$ sampled from the data distribution~$p(x,y)$; this is the model that we may quantize later in Section~\ref{sec:dfq:kd}. The class~$y$ is the condition for a CGAN. We use the pre-trained teacher as a fixed discriminator in CGAN training. Then, we propose the following objective for zero-shot learning of a CGAN:
\begin{equation}\label{sec:dfq:zs-cgan:01}
\min_G\{\expect_{p(z)p(y)}[\CE(y,T(G(z,y)))]+L_{\text{BNS}}(G)\},
\end{equation}
where $\CE$ is the cross-entropy, and $y$ is one-hot encoded in $\CE$. In comparison to the minimax optimization of CGANs in \eqref{sec:dfq:cgan:01} using the training data, we omit the maximization part, since we are given a well-trained discriminator, i.e., pre-trained teacher~$T$. We also introduce an auxiliary loss~$L_{\text{BNS}}(G)$ that constrains the generator to produce synthetic samples with statistics similar to those of the original training data.

\emph{Batch-Normalization Statistics (BNS) loss}. The batch normalization layers in a pre-trained teacher store the mean and variance of the layer input, which we can utilize as a proxy to verify that the generator output is similar to the original training data. We propose the Kullback-Leibler (KL) divergence of two Gaussian distributions to match the statistics (mean and variance) stored in the batch normalization layers of the teacher (which were obtained when trained with the original data) and the empirical statistics computed at the same batch normalization layers of the teacher for the generator output.

Let $\mu(l,c)$ and $\sigma^2(l,c)$ be the mean and variance stored in batch normalization layer~$l$ of the teacher for channel~$c$, which are obtained from the original training data. Let $\hat{\mu}_G(l,c)$ and $\hat{\sigma}_G^2(l,c)$ be the corresponding mean and variance computed for the synthetic samples from $G$. The auxiliary loss~$L_{\text{BNS}}(G)$ in \eqref{sec:dfq:zs-cgan:01} is given by
\[
L_{\text{BNS}}(G)
=\sum_{l,c}\KL_{\mathcal{N}}((\hat{\mu}_G(l,c),\hat{\sigma}_G^2(l,c)),(\mu(l,c),\sigma^2(l,c))),
\]
where $\KL_{\mathcal{N}}$ is the KL divergence of two Gaussians, i.e.,
\[
\KL_{\mathcal{N}}((\hat{\mu},\hat{\sigma}^2),(\mu,\sigma^2)) \\
=\frac{(\hat{\mu}-\mu)^2+\hat{\sigma}^2}{2\sigma^2}-\log\frac{\hat{\sigma}}{\sigma}-\frac{1}{2}.
\]

\subsection{Data-Free Knowledge Distillation for Quantization}\label{sec:dfq:kd}

Knowledge distillation (KD)~\cite{hinton2015distilling} is a well-known knowledge transfer framework to train a small ``student" network under a guidance of a large pre-trained ``teacher'' model. The original idea from Hinton et al. utilizes the soft decision output of a well-trained teacher~$T$ to help to train a student~$S$, as below:
\begin{equation}\label{sec:dfq:kd:01}
\min_S
\expect_{p(x,y)}\left[(1-\lambda)\CE(y,S(x))
+\lambda\CE(T(x),S(x))\right],
\end{equation}
for $0\leq\lambda\leq1$, where we omitted the ``temperature'' parameter in the second cross-entropy term for simplicity, which can be applied to the teacher and the student logits, respectively, before obtaining the final softmax scores.

After training a conditional generator with the proposed ZS-CGAN, we use it to generate synthetic data. The synthetic data are then used for data-free KD. In particular, for network quantization, the student is set to be a quantized version of the teacher, where its weights and activations are quantized. Then, one can seamlessly apply \eqref{sec:dfq:kd:01} to train the quantized student. In KD with the synthetic samples from ZS-CGAN, we suggest using only the second cross-entropy term in \eqref{sec:dfq:kd:01}, since the ground-truth labels for the synthetic samples are unknown although they are machine-labeled in ZS-CGAN. So we set $\lambda=1$ and perform
\begin{equation}\label{sec:dfq:kd:02}
\min_S
\expect_{p(z)p(y)}\left[\CE(T(G(z,y)),S(G(z,y)))\right].
\end{equation}

Our proposed method is considered a novel approach to data-free quantization that uses the synthetic data generated from ZS-CGAN, to train a quantized student by knowledge distillation from a floating-point teacher. It is different from the previous data-free quantization method in \cite{nagel2019data} that uses a series of transformations to re-distribute and equalize weights across layers so they have smaller deviation in each layer and suffer less from quantization. Our approach is also different from the model inversion techniques\cite{cai2020zeroq,yin2020dreaming,haroush2020knowledge} that do not train any generators. The conditional data generation by ZS-CGAN yields more distinct synthetic samples per class than those generated by non-conditional generators as in \cite{choi2020data}, and hence results in more accurate data-free network quantization.

\section{EXPERIMENTS}\label{sec:exp}

We perform experiments on the proposed ZS-CGAN for data-free network quantization. TensorFlow provides two network quantization schemes, and we use them as the baseline quantization methods. 
\begin{itemize}[noitemsep,topsep=0em,leftmargin=1.5em]
\item First, TensorFlow Lite provides $8$-bit post-training quantization (PTQ)\footnote{https://www.tensorflow.org/lite/performance/post\_training\_quantization} technique. No data are needed for weight quantization. The quantization of activations is based on some calibration data, which are used to collect the range (the minimum and the maximum) of activations and to determine the quantization bin size based on the range.
\item Second, TensorFlow's quantization-aware training (QAT)\footnote{https://github.com/tensorflow/tensorflow/tree/r1.15/tensorflow/contrib/quantize} can be used to reduce the quantization loss by (re-)train a quantized network~\cite{jacob2018quantization}. Some training data are required for QAT, and the original training dataset is usually used.
\end{itemize}
For data-free PTQ, we first train a conditional generator via the proposed ZS-CGAN with no original training data, given a pre-trained model. Then, we produce $10$k synthetic samples and use them as the calibration data in PTQ. For data-free QAT, we implement data-free KD using ZS-CGAN (see \eqref{sec:dfq:kd:02} in Section~\ref{sec:dfq:kd}) on top of TensorFlow's QAT framework.

We obtain pre-trained models\footnote{https://github.com/osmr/imgclsmob} of ResNet-18, ResNet-50, MobileNetV1, MobileNetV2, and MobileNetV3-Large that achieve accuracies of 71.92\%, 77.86\%, 73.57\%, 73.24\%, and 75.31\%, respectively, on the ImageNet 2012 validation dataset~\cite{russakovsky2015imagenet}. For the conditional generator in ZS-CGAN, we adopt the U-Net architecture used in \cite{ulyanov2018deep} and replace batch normalization with conditional batch normalization~\cite{de2017modulating,miyato2018cgans} for conditional generation.

In ZS-CGAN, we use Adam optimizer~\cite{kingma2014adam} with momentum~$0.5$ and learning rate~$10^{-3}$. In data-free KD for data-free QAT, we use Nesterov accelerated gradient with momentum~$0.9$ and learning rate~$10^{-4}$. The learning rate is annealed by cosine decaying~\cite{loshchilov2016sgdr}. We use $200$ epochs for ZS-CGAN and then $50$ epochs for data-free KD, where each epoch consists of $1000$ batches of batch size $128$. In each batch of data-free KD, we generate new synthetic samples, while we keep updating the conditional generator with ZS-CGAN.

\setlength{\tabcolsep}{.2em}
\begin{table}[t]
\centering
\caption{Data-free quantization (DF-Q) results using our ZS-CGAN. We show the results for data-free PTQ (DF-PTQ) and data-free QAT (DF-QAT), both with ZS-CGAN. We compare our method with the existing DF-Q schemes in \cite{nagel2019data,cai2020zeroq,yin2020dreaming,choi2020data}, and bold numbers indicate the best results among DF-Q schemes. For data-dependent quantization (DD-Q), we show the results for PTQ and QAT by using the original ImageNet dataset~\cite{jacob2018quantization}.\label{sec:exp:tbl:01}}\vspace{-.5em}
(a) INT8 quantization\vspace{.2em}\\
{\footnotesize
\begin{tabular}{ccccccccc}
\toprule
Model       & \multicolumn{8}{c}{Quantized model accuracy (\%)}\\
            & \multicolumn{6}{c}{DF-Q}
                                                             & \multicolumn{2}{c}{DD-Q*}\\
\cmidrule(l{2pt}r{2pt}){2-7} \cmidrule(l{2pt}r{2pt}){8-9}
            & \multicolumn{2}{c}{ZS-CGAN (ours)}
                                     & \cite{nagel2019data}
                                             & \cite{cai2020zeroq}
                                                     & \cite{yin2020dreaming}
                                                             & \cite{choi2020data}
                                                                     & \multicolumn{2}{c}{\cite{jacob2018quantization}}\\
            & DF-PTQ & DF-QAT        &       &       &       &       & PTQ   & QAT\\
\midrule
ResNet-18   & 70.27 & \textbf{71.72} & 69.70 & 71.43 & 71.35 & 71.56 & 71.00 & 71.82\\
ResNet-50   & 76.76 & \textbf{77.87} & -     & 77.67 & 77.58 & 77.81 & 77.34 & 77.86\\
MobileNetV1 & 69.20 & \textbf{73.37} & 70.50 & -     & -     & 73.06 & 72.01 & 73.54\\
MobileNetV2 & 72.37 & \textbf{73.05} & 71.20 & 72.91 & 72.45 & 72.91 & 72.58 & 73.24\\
MobileNetV3-L & 73.45 & \textbf{74.75} & -     & -     & 74.41 & 74.71 & 73.64 & 74.97\\
\bottomrule
\multicolumn{9}{r}{* Used the original ImageNet dataset.}
\end{tabular}
}\\\vspace{.2em}
(b) INT6 quantization\vspace{.2em}\\
{\footnotesize
\begin{tabular}{ccccc}
\toprule
Model       & \multicolumn{4}{c}{Quantized model accuracy (\%)}\\
            & \multicolumn{3}{c}{DF-QAT}
                                             & DD-QAT*\\
\cmidrule(l{2pt}r{2pt}){2-4} \cmidrule(l{2pt}r{2pt}){5-5}
            & ZS-CGAN (ours)
                             & \cite{yin2020dreaming}
                                     & \cite{choi2020data}
                                             & \cite{jacob2018quantization}\\
\midrule
ResNet-18   & \textbf{70.78} & 70.20 & 70.73 & 71.02\\
ResNet-50   & \textbf{76.82} & 76.47 & 76.44 & 77.06\\
MobileNetV1 & \textbf{71.05} & -     & 70.81 & 72.01\\
MobileNetV2 & \textbf{72.24} & 71.47 & 71.93 & 72.51\\
\bottomrule
\end{tabular}\vspace{-1.2em}
}
\end{table}

We show our data-free quantization results (DF-PTQ and DF-QAT) using the synthetic data from ZS-CGAN in Table~\ref{sec:exp:tbl:01}. Both weights and activations are quantized into $8$-bit integers for INT8 models and $6$-bit integers for INT6 models, respectively. We compare our method to the existing data-free quantization (DF-Q) schemes in \cite{nagel2019data,cai2020zeroq,yin2020dreaming,choi2020data}. Although DF-Q was not evaluated in \cite{yin2020dreaming}, we produced the DF-Q results of \cite{yin2020dreaming} by performing QAT with $1$M synthetic samples generated from their model inversion method\footnote{https://github.com/NVlabs/DeepInversion}. We also implemented the DF-QAT method in \cite{choi2020data}, which uses a non-conditional generator. 
The proposed scheme achieves the state-of-the-art data-free quantization results for all ResNet and MobileNet models trained on the ImageNet dataset. Observe that using the conditional generator from ZS-CGAN outperforms using the non-conditional generator from \cite{choi2020data}. Table~\ref{sec:exp:tbl:01} also includes the data-dependent quantization (DD-Q) results using the original ImageNet training data. The accuracy loss from using the original ImageNet dataset is very marginal.

\setlength{\tabcolsep}{0.3em}
\begin{table}[t]
\centering
\caption{Comparison of DF-QAT when ZS-CGAN is trained with the cross-entropy (CE) loss and/or the loss of matching batch-normalization statistics (BNS) (see \eqref{sec:dfq:zs-cgan:01}). We also show the QAT results with a pre-trained BigGAN model.\label{sec:exp:tbl:03}}\vspace{-.5em}
{\footnotesize
\begin{tabular}{lccc}
\toprule
Data                              & Weight & Activation & ResNet-18\\
                                  &        &            & accuracy (\%)\\
\midrule
Pre-trained with original ImageNet
                                  & FL32   & FL32       & 71.92\\
\midrule
Synthetic from ZS-CGAN (CE+BNS)   & INT8   & INT8       & \textbf{71.72}\\
Synthetic from ZS-CGAN (CE)       & INT8   & INT8       & 69.63\\
Synthetic from ZS-CGAN (BNS)      & INT8   & INT8       & 71.69\\
\midrule
Synthetic from BigGAN~\cite{brock2018large}
                                  & INT8   & INT8       & 71.69\\
Original ImageNet
                                  & INT8   & INT8       & 71.82\\
\bottomrule
\end{tabular}\vspace{-1.2em}
}
\end{table}

For ablation study, we compare ZS-CGAN with and without each loss term of \eqref{sec:dfq:zs-cgan:01} in Table~\ref{sec:exp:tbl:03}. The best data-free QAT result is obtained by using both the cross-entropy (CE) loss and the loss to match the batch-normalization statistics (BNS) in ZS-CGAN. We also perform QAT by using the synthetic samples from a pre-trained BigGAN model\footnote{https://github.com/huggingface/pytorch-pretrained-BigGAN}. BigGAN~\cite{brock2018large} is one of the state-of-the-art CGANs for ImageNet. The original ImageNet data are used to train the BigGAN model, and so it is not data-free; recall that our ZS-CGAN does not use any ImageNet data in training its conditional generator. The synthetic samples from ZS-CGAN yield comparable QAT performance to the synthetic samples from the pre-trained BigGAN.

\setlength{\tabcolsep}{0.1em}
\begin{figure}[!t]
\centering
{\scriptsize
\begin{tabular}{cccccc}
hen & frog & retriever & bear & gorilla & dam\\
\includegraphics[width=.076\textwidth]{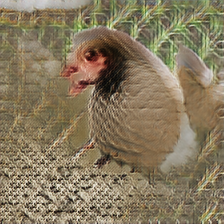}&
\includegraphics[width=.076\textwidth]{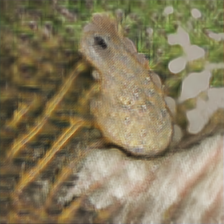}&
\includegraphics[width=.076\textwidth]{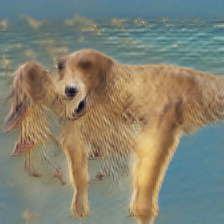}&
\includegraphics[width=.076\textwidth]{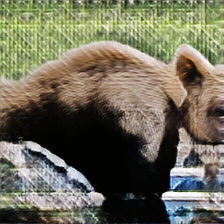}&
\includegraphics[width=.076\textwidth]{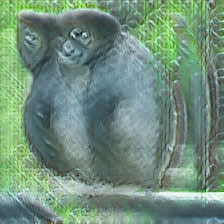}&
\includegraphics[width=.076\textwidth]{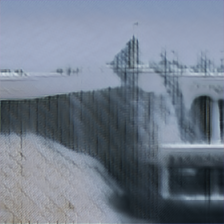}\\
monastery & pirate ship & pitcher & wall clock & wine bottle & pretzel\\
\includegraphics[width=.076\textwidth]{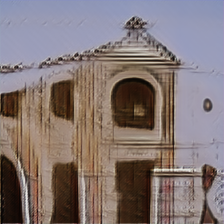}&
\includegraphics[width=.076\textwidth]{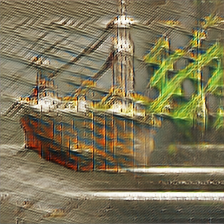}&
\includegraphics[width=.076\textwidth]{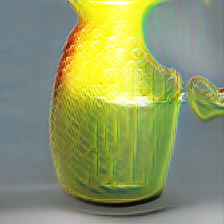}&
\includegraphics[width=.076\textwidth]{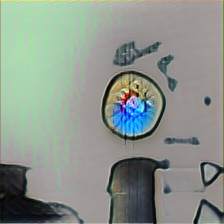}&
\includegraphics[width=.076\textwidth]{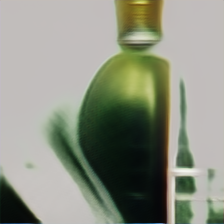}&
\includegraphics[width=.076\textwidth]{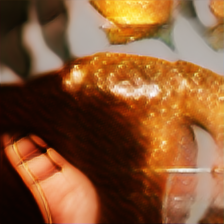}\\
\end{tabular}
}
(a) ZS-CGAN trained with both CE and BNS losses\\\vspace{.2em}
{\scriptsize
\begin{tabular}{cccccc}
hen & frog & retriever & bear & gorilla & dam\\
\includegraphics[width=.076\textwidth]{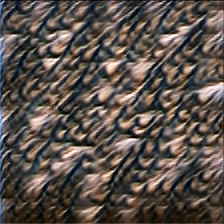}&
\includegraphics[width=.076\textwidth]{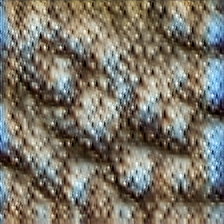}&
\includegraphics[width=.076\textwidth]{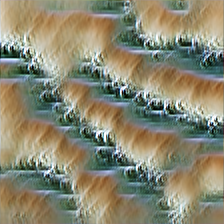}&
\includegraphics[width=.076\textwidth]{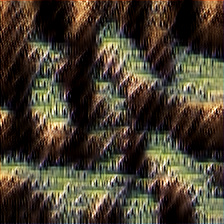}&
\includegraphics[width=.076\textwidth]{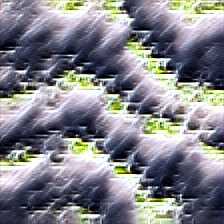}&
\includegraphics[width=.076\textwidth]{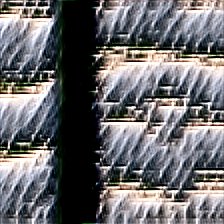}\\
monastery & pirate ship & pitcher & wall clock & wine bottle & pretzel\\
\includegraphics[width=.076\textwidth]{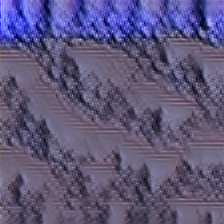}&
\includegraphics[width=.076\textwidth]{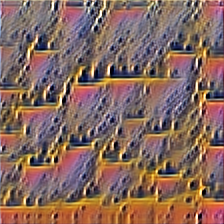}&
\includegraphics[width=.076\textwidth]{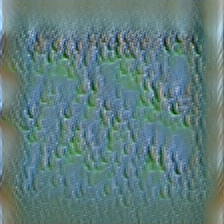}&
\includegraphics[width=.076\textwidth]{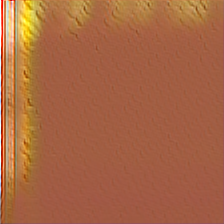}&
\includegraphics[width=.076\textwidth]{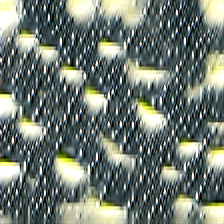}&
\includegraphics[width=.076\textwidth]{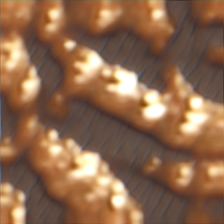}\\
\end{tabular}
}
(b) ZS-CGAN trained with CE loss only\\\vspace{.2em}
{\scriptsize
\begin{tabular}{cccccc}
\includegraphics[width=.076\textwidth]{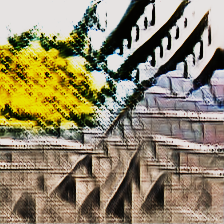}&
\includegraphics[width=.076\textwidth]{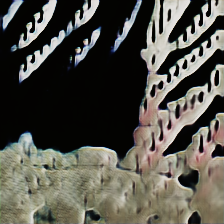}&
\includegraphics[width=.076\textwidth]{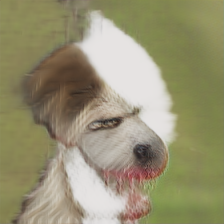}&
\includegraphics[width=.076\textwidth]{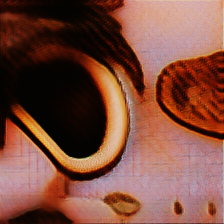}&
\includegraphics[width=.076\textwidth]{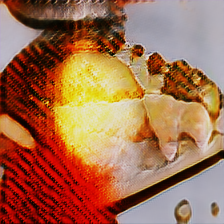}&
\includegraphics[width=.076\textwidth]{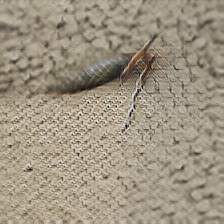}\\
\includegraphics[width=.076\textwidth]{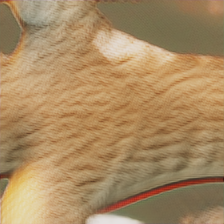}&
\includegraphics[width=.076\textwidth]{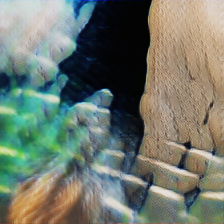}&
\includegraphics[width=.076\textwidth]{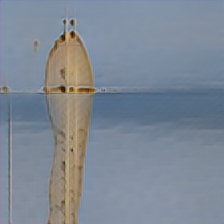}&
\includegraphics[width=.076\textwidth]{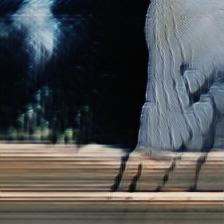}&
\includegraphics[width=.076\textwidth]{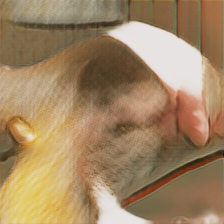}&
\includegraphics[width=.076\textwidth]{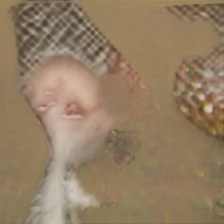}\\
\end{tabular}
}
(c) ZS-CGAN trained with BNS loss only\\\vspace{-.5em}
\caption{Exemplary synthetic images from ZS-CGAN, given a pre-trained ResNet-18 on the ImageNet dataset.\label{sec:dfq:cgen:fig:01}}\vspace{-1.2em}
\end{figure}

In Figure~\ref{sec:dfq:cgen:fig:01}, we show exemplary synthetic samples from ZS-CGAN, given the pre-trained ResNet-18 model on the ImageNet dataset as the teacher. When we use both CE and BNS losses, ZS-CGAN produces labeled images that we can distinguish their classes. When we use the CE loss only, the conditional generator fails to mimic natural images of given classes. If we use the BNS loss only, then no feedback (critic) for the label of a generated sample is delivered from the pre-trained model (discriminator) to the conditional generator, and so it produces synthetic samples without labels, while just matching the statistics at the batch normalization layers.

\section{CONCLUSION}\label{sec:con}

In this paper, we proposed zero-shot learning of a conditional generator for data-free network quantization. No original data are used in the proposed pipeline for training the conditional generator and for quantization of a pre-trained model. A pre-trained classification model is used as a fixed discriminator in CGAN training. In particular, we proposed using the statistics at the batch normalization layers of the pre-trained model to additionally constrain the generator to produce samples similar to the original training data. For data-free network quantization, we obtained quantized models that achieve comparable accuracy to the models quantized with the original training dataset. Finally, it remains as our future work to apply the proposed framework to other network compression tasks such as network pruning and neural architecture search.


{
\small
\newcommand{\BIBdecl}{\setlength{\itemsep}{0.25em}}
\bibliographystyle{IEEEbib}
\bibliography{ref}
}

\end{document}